\documentclass[sigconf, nonacm]{acmart}

\AtBeginDocument{%
  \providecommand\BibTeX{{%
    \normalfont B\kern-0.5em{\scshape i\kern-0.25em b}\kern-0.8em\TeX}}}

\usepackage{textcomp}
\usepackage{bbm}
\usepackage{makecell}
\usepackage{array}
\usepackage{multirow}
\usepackage{algorithm}
\usepackage{algpseudocode}
\usepackage{caption}
\usepackage{subcaption}

\begin{document}

\title{ActiveDP: Bridging Active Learning and Data Programming}

\author{Naiqing Guan}
\affiliation{%
  \institution{University of Toronto}
  \city{Toronto}
  \country{Canada}
}
\email{naiqing.guan@mail.utoronto.ca}

\author{Nick Koudas}
\affiliation{%
  \institution{University of Toronto}
  \city{Toronto}
  \country{Canada}
}
\email{koudas@cs.toronto.edu}


\begin{abstract}
  Modern machine learning models require large labelled datasets to achieve good performance, but manually labelling large datasets is expensive and time-consuming. The data programming paradigm enables users to label large datasets efficiently but produces noisy labels, which deteriorates the downstream model's performance. The active learning paradigm, on the other hand, can acquire accurate labels but only for a small fraction of instances. In this paper, we propose ActiveDP, an interactive framework bridging active learning and data programming together to generate labels with both high accuracy and coverage, combining the strengths of both paradigms. Experiments show that ActiveDP outperforms previous weak supervision and active learning approaches and consistently performs well under different labelling budgets.
\end{abstract}

\ccsdesc[500]{Information systems~Data cleaning}
\ccsdesc[500]{Computing methodologies~Active learning settings}

\keywords{Dataset Curation, Data programming, Active learning}



\maketitle

\section{Introduction} \label{sec:intro}
Modern machine learning models require large training datasets to achieve good accuracy, yet manual labelling and curation of large datasets are both expensive and time-consuming. Thus, acquiring labelled datasets has become one of the main bottlenecks in applying machine learning in practical scenarios. 
Data programming (DP) \cite{ratner2016data,ratner2017snorkel}, a recent paradigm for weak supervision, provides an approach to automatically label large datasets without manually annotating specific instances. In the data programming paradigm, users represent weak supervision sources in the form of label functions (LFs), which are rules that provide noisy labels to a subset of data. For example, suppose the task is to classify the sentiment of customer reviews. In that case, the users can write code snippets that label the review as positive or negative based on specific keywords. Since the label functions have varying accuracy and may exhibit ad-hoc correlations, a generative model (also called the label model) is designed to aggregate noisy, weak labels into probabilistic labels. Optionally, a downstream model (also referred to as the end model) can be trained with the probabilistic labels and utilized for downstream tasks. 

While the data programming paradigm can label large datasets rapidly, the generated labels are usually noisy, and as a result this may deteriorate the performance of the downstream model. Active learning (AL) \cite{settles2009active,settles2012active}, on the other hand, identifies and selects a subset of data for manual annotation and utilizes the subset of data for training the downstream model. As the annotation is instance-specific, users can provide cleaner labels to the selected instances. While active learning excels in label quality, it falls short in label quantity, as users can only annotate a small fraction of data with limited effort. 

\begin{table}[t]
\footnotesize
\centering
\caption{Approaches briding active learning with data programming.}\label{tab:approaches}
\vspace{-0.1cm}
\resizebox{\columnwidth}{!}{
\begin{tabular}{c c c c}
\toprule[1.5pt]
\textbf{Method} & \textbf{Query Type} & \textbf{Train AL Model?} & \textbf{Step to Improve Data Quality}\\
\hline
Nemo \cite{hsieh2022nemo} & LF Creation & No & LF Generation \\
\hline
IWS \cite{boecking2020interactive} & LF Validation & No & LF Generation \\
\hline
Revising LF \cite{nashaat2018hybridization} &  Instance Labelling & No &  LF Revision \\
\hline
Active WeaSuL \cite{biegel2021active} & Instance Labelling & No & Label Model Training\\
\hline
ActiveDP (ours) & LF Creation & Yes & Post Training \\
\bottomrule[1.5pt]
\end{tabular}}
\end{table}

It is evident that the merits of data programming and active learning compensate for each other. In this paper, we address the question: \textit{how can we combine the advantages of both methods to label large datasets rapidly while maintaining superior label quality? } Researchers have investigated several approaches to bridge data programming and active learning, such as leveraging active learning to develop label functions \cite{hsieh2022nemo, boecking2020interactive}, revise label functions \cite{nashaat2018hybridization} or tune label model parameters \cite{biegel2021active}. However, these methods do not fully leverage the merits of both paradigms. Nemo \cite{hsieh2022nemo} and IWS \cite{boecking2020interactive} leverage active learning to select instances for LF development. However, they only use LFs to annotate the training dataset instead of combining the LFs with fine-grained instance labels, which impedes them from achieving high-quality labels. Nashaat et al. \cite{nashaat2018hybridization} correct the LF outputs on labelled data. However, as pointed out by Biegel et al. \cite{biegel2021active}, the revision does not always benefit predictions on unlabelled data, as revising the LFs on the labelled subset may mislead the label model to believe they are also accurate on the unlabeled subset. Biegel et al. \cite{biegel2021active} use the labelled subset to tune label model parameters. However, it only leverages label functions to generate training labels and thus has a similar drawback to Nemo and IWS. Besides, the trade-off factor between the original model loss and the loss term induced on the labelled subset is hard to select. 

\begin{figure*}[t]
\centering
 \includegraphics[width=2\columnwidth]{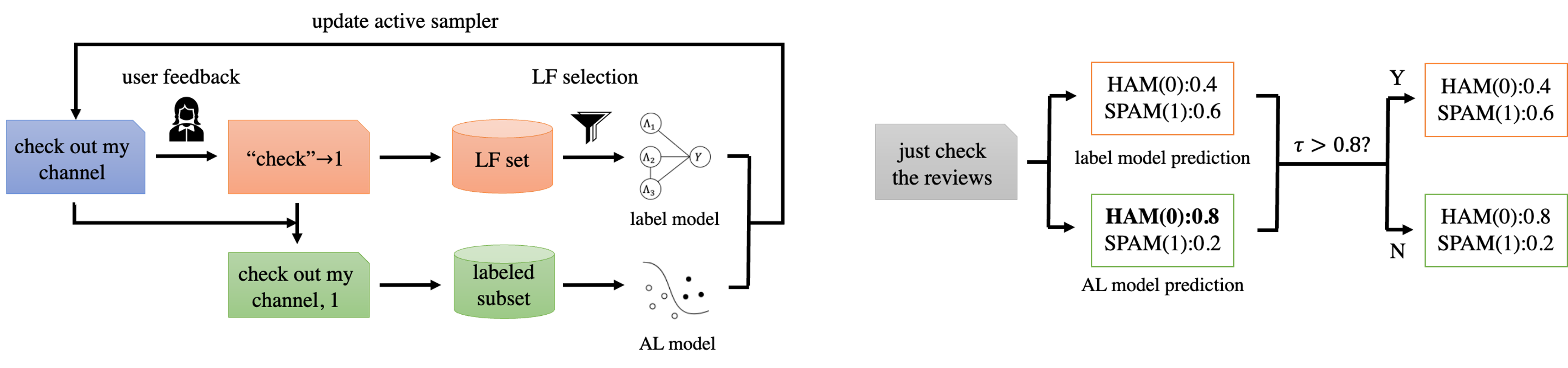}
 \caption{Workflow of ActiveDP. Left: iterative LF creation at training phase. Right: label aggregation at inference phase.}\label{fig:activeDP-workflow}
\end{figure*}

To address the abovementioned limitations, we propose ActiveDP, a novel interactive framework bridging active learning and data programming.  Table \ref{tab:approaches} highlights the difference between ActiveDP and previous works \cite{nashaat2018hybridization,biegel2021active,boecking2020interactive,hsieh2022nemo}. The main differences are: (1) ActiveDP combines weak supervision with active learning model predictions to balance label quality and quantity; (2) ActiveDP improves the label quality in the post-training phase. This enables ActiveDP to treat the label model as a black box and neglect the intricate effect of revising LFs on label model training.

Figure \ref{fig:activeDP-workflow} illustrates the workflow of the ActiveDP framework. In the training phase, ActiveDP leverages a novel ADP sampler to select instances for users to develop LFs. Apart from collecting the LFs returned by users, ActiveDP also creates a small pseudo-labelled dataset based on user-returned LFs to train an active learning model. Besides, ActiveDP learns the dependency structure between LFs and the labels and chooses a diverse and informative set of LFs to improve the end-to-end classification accuracy. In the inference phase, ActiveDP aggregates the prediction results of the label model in the DP paradigm and the active learning model using a novel label aggregation method named {\em ConFusion} to generate labels with high accuracy and coverage. 

To summarize, we make the following main contributions:
\begin{itemize}

\item We propose a novel interactive framework named ActiveDP that explores the design space between active learning and data programming and combines their strengths.

\item We design various novel strategies to improve the efficiency of ActiveDP and the quality of generated labels, including the ConFusion method for label aggregation, the ADP sampler for query instance selection, and the LabelPick method for LF filtering.




\item We conduct extensive experiments to compare ActiveDP with baseline methods, demonstrating its efficiency in providing labels with high coverage and accuracy and enhancing the performance of the downstream model.
\end{itemize}

\section{Related Works} \label{sec:background}

\subsection{Data Programming}
Data programming (DP) is a weak supervision paradigm for automatically labelling large datasets. Given an unlabeled dataset $D=\{x_i\}_{i=1}^n$, the users design a series of label functions $\Lambda=\{\lambda_j\}_{j=1}^m$, each provide weak labels to a subset of data. We use $\lambda_j(x_i)$ to denote the output of the $\lambda_j$ on $x_i$. For a class classification problem with $Y=\{0,1,...,C-1\}$ as the label space, the weak label $\lambda_j(x_i)\in Y \cup \{-1\}$ where $-1$ means $\lambda_j$ does not make prediction on $x_i$. We say $\lambda_j$ is activated on $x_i$ if $\lambda_j(x_i)\neq -1$ and $\lambda_j$ abstains on $x_i$ otherwise. The generated weak labels form a label matrix $W$ where $W_{ij} = \lambda_j(x_i)$, which is used to train a label model $f_l$. We use $f_l(x,\Lambda)\in R^C$ to denote the probabilistic labels predicted by the label model for $x$. Finally, the probabilistic labels (or hard labels by selecting the most likely class) will be used to train the downstream machine learning model.

 The works in the DP area can be briefly categorized into developing more advanced label models \cite{ratner2016data,ratner2017snorkel,ratner2019training,awasthi2019learning,biegel2021active} and designing label functions efficiently \cite{varma2018snuba,boecking2020interactive, hsieh2022nemo, denham2022witan, hancock2018training,smith2022language,huangscriptoriumws}. 
Snorkel \cite{ratner2016data,ratner2017snorkel} uses a factor graph to model the accuracy and pairwise dependencies between label functions. MeTaL \cite{ratner2019training} extends the framework to multi-task settings and proposes to estimate LF accuracy by solving a matrix completion problem. 
Another line of work focuses on automating or aiding LF design.  IWS \cite{boecking2020interactive} incorporates human experts in an iterative process by letting experts decide whether a candidate LF is accurate. Nemo \cite{hsieh2022nemo} actively selects the instances given to human experts, and experts return LFs based on them.

\subsection{Active Learning}
Active learning \cite{settles2009active,settles2012active} is an established technique in machine learning literature, which aims to maximize model accuracy while annotating the fewest samples possible. There are abundant selectors designed for active learning, such as uncertainty sampling \cite{lewis1995sequential}, query-by-committee \cite{seung1992query}, core-set approaches \cite{sener2018active}, density-based approaches \cite{settles2008analysis} and hybrid approaches \cite{li2013adaptive,gao2020consistency}. We refer interested readers to recent surveys \cite{ren2021survey,liu2022survey} for a comprehensive description of active learning methods.


Recently, there have been emerging works that leverage active learning
to improve the performance of PWS pipelines. Nashaat et al. \cite{nashaat2018hybridization, nashaat2020asterisk} leverage uncertainty sampling to correct LF outputs. Active WeaSuL \cite{biegel2021active} actively selects a subset of instances for labelling to guide label model training. IWS \cite{boecking2020interactive} and Nemo \cite{hsieh2022nemo} also leverages active learning for sample selection.

\section{ActiveDP Framework} \label{sec:method}
In this section, we introduce and design the ActiveDP framework. We first provide an overview of the ActiveDP framework and illustrate its workflow using a running example; then, we discuss each component of ActiveDP in detail. 

\subsection{Framework Overview}
As illustrated in Figure 
\ref{fig:activeDP-workflow}, the ActiveDP framework contains a training phase and an inference phase. In the training phase, it iteractively selects query instances for users to inspect and design LFs, and trains the label model and active learning model based on user responses. In the inference phase, it aggregates the prediction results of the label model and the active learning model to improve label quality.

In the training phase, during the $t$-th iteration, the system leverages the ADP Sampler (Section \ref{sec:sample}) to actively pick an example $x_{l_t}\in D$ as the query instance for the user to inspect. The user designs a LF $\lambda_t$ based on $x_{l_t}$. In the running example in Figure \ref{fig:activeDP-workflow} for spam detection, the sampler picks query instance \textit{"check out my channel"} and asks the user to suggest a LF based on the query, and the user responds with a LF \textit{"check"$\to$ SPAM(1)}, indicating that a comment is likely to be spam when it contains the keyword "check". The LFs will be collected into the LF set $\Lambda_t=\{\lambda_1,...,\lambda_t\}$. We further leverage the LabelPick method (Section \ref{sec:lf-selection}) to select a subset of helpful LFs $\Lambda_t^* \subset \Lambda_t$ , and use $\Lambda_t^*$ to train the label model $f_l^t$.

Apart from collecting the LFs, we also curate a pseudo-labelled subset of the training data to train an active learning model. While the user does not explicitly provide labels to data, since we have the LFs designed by the user based on the query instances, these LFs should be at least accurate on the corresponding query instances. In the running example, as the user designs a LF \textit{"check"$\to$ SPAM(1)} when they observe the query instance \textit{"check out my channel"}, we can infer the label of the query instance should be SPAM. Based on that observation, we create the pseudo-labelled subset as $L_t=\{(x_{l_t}, \tilde{y}_{l_t})\}_{t=1}^T$, where $\tilde{y}_{l_t}=\lambda_t(x_{l_t})$ being the pseudo-label inferred from LFs, and use $L_t$ to train the active learning model $f_a^t$. 

In the inference phase, we leverage the ConFusion method (Section \ref{sec:label-agg}) to aggregate the prediction results of the label model and active learning model based on a confidence threshold $\tau_t$. We would follow the prediction of the active learning model if its confidence exceeds the confidence threshold. Otherwise, we would follow the prediction of the label model as long as the instance has non-abstain LFs. In the running example in Figure \ref{fig:activeDP-workflow}, the unlabeled instance is \textit{"just check the reviews"}. Suppose the active learning model predicts it to be HAM with confidence score 0.8, and the confidence threshold is 0.7. Since the confidence score exceeds the threshold, we will follow the active learning model's prediction for that unlabeled instance.

\subsection{Label Aggregation}\label{sec:label-agg}
In this section, we introduce ConFusion, a confidence-based label aggregation approach.  While the label model and the active learning model are both trained on the training dataset, their predictions differ because they are trained using different features and their model structures are different. This makes label aggregation helpful in combining the strengths of both models and improves label quality. We use $f_l^t(x,\Lambda^*_t)$ and $f_a^t(x)$ to denote the soft labels predicted by the label model and the active learning model after the t-th iteration respectively. The confidence score of the active learning model is defined as the predicted probability for the top-1 candidate class, i.e. $max\{f_a^t(x)\}$. For example, the confidence score for the unlabeled instance in Figure \ref{fig:activeDP-workflow} is 0.8. The aggregated labels are defined as follows:
\begin{equation}
\hat{y}^t = 
\begin{cases}
f_a^t(x) & max \{f_a^t(x)\} \geq \tau_t \\
f_l^t(x,\Lambda_t^*) & max \{f_a^t(x)\} < \tau_t \wedge \Lambda_t^*(x) \neq \mathbf{-1} \\
\emptyset & otherwise.
\end{cases}
\label{eq:agg}
\end{equation}

 In other words, when the confidence of the active learning model is above the threshold $\tau_t$, we adopt its prediction results. Otherwise, if at least one label function is activated on the unlabeled instance (i.e. $\exists \lambda \in \Lambda_t^*\ s.t. \lambda(x)\neq -1$), we adopt the prediction results of the label model. If all label functions abstain on the unlabeled instance and the confidence of the active learning model is below $\tau_t$, we reject making predictions and discard the instance when training the downstream model.  Such an approach balances label accuracy and label coverage because we leverage data programming to label a significant fraction of data while using the active learning model to improve label accuracy. 

The confidence threshold $\tau_t$ determines the relative importance of the active learning model and the label model.  We dynamically tune the confidence threshold $\tau_t$ using a holdout validation set $D_V$ drawn I.I.D. from the underlying distribution of training data (with hidden labels). Let $C^t=\{0.0,c_1^t,c_2^t,...,c^t_k,1.0\}$ be the unique confidence scores of the active learning model $f_a^t$ on instances in $D_V$ together with two boundary values 0.0 and 1.0. We evaluate every candidate threshold in $C^t$, use Equation \ref{eq:agg} to aggregate labels and compute the accuracy of the aggregated labels on the validation set. Note that when we evaluate the label accuracy, we only consider the part of validation data not rejected by the ConFusion method due to low confidence. Finally, we choose $\tau_t\in C^t$ to maximize the aggregated label's accuracy on the validation set.

Both high label accuracy and coverage are desirable for training the downstream models, thus it is also possible to select $\tau_t\in C^t$ that maximize the aggregated label's coverage on the validation set following a similar process. However, selecting $\tau_t = 0.0$ would always lead to maximal coverage as the active learning model would be active on every instance, making ActiveDP fall back to active learning. In our experiments, we observe that improving label accuracy is more important in enhancing the end model performance than improving label coverage, which also motivates us to maximize label accuracy during threshold tuning.

\subsection{Sample Selection Strategy} \label{sec:sample}
As our framework combines the prediction results of the active learning model and the label model, the sample selector we design needs to balance between two goals: improving the performance of the active learning model and guiding experts to design helpful LFs. To trade-off between these goals, we use entropy \cite{shannon1948mathematical} to evaluate the uncertainty of both models, and propose the ADP sampler that selects the most uncertain point based on the prediction of both models:
\begin{equation}
x_{t+1}^* = argmax_x[Ent(f_a^t(x))^\alpha * Ent(f_l^t(x,\Lambda_t^*))^{1-\alpha}]
\end{equation}

Where $Ent()$ is the entropy function defined as follows. 
\begin{equation}
Ent(p)= -\sum_i{p_i*log(p_i)}
\end{equation}

The trade-off factor $\alpha\in[0,1]$ balances between the active learning model and the label model. We set $\alpha=0.5$ for textual datasets and $\alpha=0.99$ for tabular datasets in our experiments. We give higher weights to the active learning model in tabular datasets because these datasets are relatively easy to classify, thus the active learning model can reach good performance with a small labelling budget.

\subsection{Label Function Selection} \label{sec:lf-selection}

\begin{figure}[t]
\centering
 \includegraphics[width=\columnwidth]{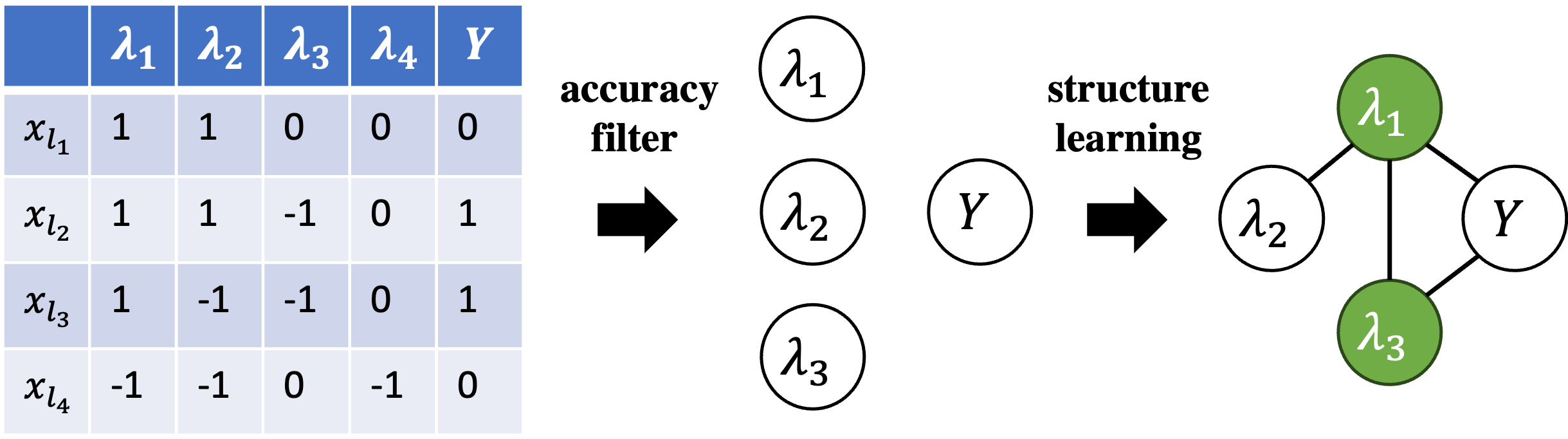}
 \caption{Workflow of the LF selection module in ActiveDP.}\label{fig:lf-selection}
\end{figure}

In this section, we introduce the LabelPick method for LF selection, which selects a subset of helpful LFs $\Lambda_t^*\subset \Lambda_t$ to train the label model. Figure \ref{fig:lf-selection} illustrates the LabelPick method. The main contribution of the LabelPick method is reducting the LF selection problem into feature selection in supervised settings. In LabelPick, we first evaluate the accuracy of candidate LFs using a holdout validation set $D_V$ drawn I.I.D. from the underlying distribution of training data (with hidden labels) and prune out LFs performing worse than random on the validation set; then we create a small labelled dataset $L_\Lambda^t = \{(\Lambda_t(x_{l_i}), \tilde{y}_{l_i}\}_{i=1}^t$. The table in Figure \ref{fig:lf-selection} illustrates the labelled dataset $L_\Lambda^t$ we created, which consists the weak labels for four query instances and their pseudo labels inferred from user feedback. We leverage the graphical lasso method \cite{friedman2008sparse} to infer the dependency structure between the LFs and the label based on $L_\Lambda^t$, and select the subset of LFs adjacent to the class label in the dependency structure, corresponding to choosing the Markov Blanket \cite{pearl1988probabilistic} of the class label. Formally, if a subset of LFs $\Lambda_t^* \subset \Lambda_t$ forms the Markov Blanket of label $Y$, then we have $Y \perp\!\!\!\perp \Lambda_t/\Lambda_t^*|\Lambda_t^*$. In other words, the Markov Blanket contains all the helpful information for inferrring the label, making other LFs unecessary once the LFs corresponding to the Markov Blanket are selected. Therefore, we can prune out the other LFs without sacrificing the label model's performance.Figure \ref{fig:lf-selection} shows that in the running example, the LFs $\lambda_1$ and $\lambda_3$ forms the Markov Blanket of $Y$, thus we select these two LFs and discard $\lambda_2$ as it is redundant given $\lambda_1$ and $\lambda_3$. $\lambda_4$ was discarded in the previous step due to low accuracy.

\section{Experiments} \label{sec:exp}
In this section, we evaluate the ActiveDP framework using the classification performance of the downstream model, compare it with other interactive frameworks for data labelling, and conduct extensive studies to investigate the benefits of each component. 

\subsection{Experiment Setup} \label{sec:exp-setup}
\subsubsection{Datasets.} 
We conduct experiments with six textual datasets that have been used to evaluate previous data programming works \cite{hsieh2022nemo,boecking2020interactive}: Youtube Spam \cite{alberto2015tubespam}, IMDB Review \cite{maas2011learning}, Amazon Review \cite{he2016ups}, Yelp Review \cite{zhang2015character}, and two subsets of the BiasBios dataset \cite{de2019bias} that distinguish between professor and teacher (marked as Bios-PT), and between journalist and photographer (marked as Bios-JP) respectively. These datasets cover three tasks: spam classification, sentiment analysis and biography classification. Besides, we also evaluated our framework on two tabular datasets: Occupancy \cite{candanedo2016accurate}, aiming predicting office room occupancy, which has been used to evaluate the Revising LF baseline \cite{nashaat2018hybridization}; Census \cite{kohavi1996scaling}, which aims at predicting whether the income of a person exceeds 50K or not.
For each dataset, we randomly partition the data into 80\%/10\%/10\% for the train-validation-test split. The detailed information of the datasets is listed in Table \ref{tab:datasets}.

\begin{table}[t]
\footnotesize
\centering
\caption{Datasets used in Evaluation.}\label{tab:datasets}
\vspace{-0.1cm}
\resizebox{\columnwidth}{!}{
\begin{tabular}{c c c c c}
\toprule[1.5pt]
\textbf{Name} & \textbf{Task} & \textbf{\#Train} & \textbf{\#Valid} & \textbf{\#Test}\\
\hline
Youtube & Spam classification & 1,566 & 195 & 195 \\
\hline
IMDB & Sentiment analysis & 20,000 & 2,500 & 2,500 \\
\hline
Yelp & Sentiment analysis & 20,000 & 2,500 & 2,500 \\
\hline
Amazon & Sentiment analysis & 20,000 & 2,500 & 2,500 \\
\hline
Bios-PT & Biography classification & 19,672 & 2,458 & 2,458 \\
\hline
Bios-JP & Biography classification & 25,808 & 3,225 & 3,225 \\
\hline
Occupancy & Occupancy prediction & 14,317 & 1,789 & 1,789 \\
\hline
Census & Income classification & 25,541 & 3,192 & 3,192 \\
\bottomrule[1.5pt]
\end{tabular}}
\end{table}
\subsubsection{Baselines.}
We compare the ActiveDP framework with the following baseline methods.
\begin{itemize}
\item Nemo \cite{hsieh2022nemo}: Nemo\footnote{\url{https://github.com/ChengYuHsieh/Nemo}} is a representative framework of the IDP paradigm. Nemo is also the current state-of-the-art work in automatically labelling large datasets. As Nemo only supports textual datasets and its SEU strategy is designed for textual data, we compare ActiveDP with Nemo on the six textual datasets only.

\item IWS \cite{boecking2020interactive}: IWS\footnote{\url{https://github.com/benbo/interactive-weak-supervision}} actively selects a candidate label function for human verification in each iteration. We evaluate IWS under the unbounded setting (IWS-LSE-a in \cite{boecking2020interactive}), where the final LF set includes all LFs that the system predicts as accurate. 

\item Revising LF (RLF) \cite{nashaat2018hybridization}: Nashaat et al. propose to iteratively select the instances where the current label model is most uncertain for human labelling and use the labels to correct LF outputs on selected instances.

\item Uncertainty Sampling (US) \cite{lewis1995sequential}: Uncertainty Sampling is a classical method for active learning, where the system selects the data instance with the highest predictive entropy for manual labelling. 
\end{itemize}

\subsubsection{Evaluation Protocol.}
We evaluate all frameworks by simulating 300 iterations of manual annotation, assess the performance of the downstream model every ten iterations, and plot the performance curve for the downstream model. We use MeTaL \cite{ratner2019training} as the label model to aggregate the label functions. We train a logistic regression model as the active learning model in ActiveDP for label quality improvement. When training the downstream model, we extract the TF-IDF representation of the input text and trains a logistic regression model as the downstream ML model for classification. We adopt classification accuracy on the test set for the downstream model performance evaluation metric and report the average test accuracy during the run, corresponding to the area under the performance curve. We repeat each experiment 5 times using different seeds and report average results. 

As different frameworks require different human supervision types, we set each iteration corresponding to labelling a single instance in uncertainty sampling and Revising LF, verifying a single LF in IWS, and designing a single LF in Nemo and ActiveDP. While the user response time may differ in each framework, such an evaluation protocol lets us compare the frameworks intuitively. \footnote{Nemo \cite{hsieh2022nemo} provides a user study on the response time for different supervision types. Asking users to give an LF takes slightly more time for users to label a single instance or verify a given LF. } Besides, Revising LF \cite{nashaat2018hybridization} requires a pre-specified set of LFs, which is not required by other frameworks. For a fair comparison, we use the same method applied in ActiveDP to create a set of LFs during the interaction and use $\Lambda_t$ as the input LF set for Revising LF when evaluating it at the t-th iteration.

\subsubsection{Simulated User.} \label{sec:simulated-user} Following previous works \cite{hsieh2022nemo, boecking2020interactive}, we use ground truth examples to simulate user responses when evaluating all frameworks. We apply the following simulation process for ActiveDP. For textual datasets,  we first consider the candidate LF space with all LFs $\lambda_{w,y}$ that return a specific class label $y$ when observing keyword $w$ and abstain otherwise. Given a query instance $x$, we first build a candidate LF set $\Lambda^c = \{\lambda_{w,y}: w \in x \wedge acc(\lambda_{w,y})> t\}$, which contains all LFs with their keywords inside the instance $x$ and with accuracy above threshold $t$. For tabular datasets, we consider the candidate LF space with all decision stumps $\lambda_{j,v,op,y}$ that returns a specific class label $y$ for instance $x$ if $x_j\geq y (op = '\geq')$ or $x_j\leq y (op = '\leq') $, and returns $-1$ otherwise. Given a query instance $x$, we build the candidate LF set as $\Lambda^c = \{\lambda_{j,v,op,y}: 1\leq j \leq m_{feat} \wedge op \in \{'\leq','\geq'\} \wedge v=x_j \wedge acc(\lambda_{j,v,op,y})> t\}$, where $m_{feat}$ is the number of features in the dataset. In other words, we consider the decision stumps based on a single feature with $x$ lying on the boundary. We set the accuracy threshold $\tau_{acc}=0.6$ in our experiments. Then we filter out LFs returned in previous iterations and select an LF from the filtered set with probability proportional to the LF coverages. For IWS, the simulated user marks a candidate LF as accurate when its accuracy exceeds $\tau_{acc}$. For uncertainty sampling and Revising LF, the simulated user returns the correct label of the selected instance. 

\subsection{System Performance} \label{sec:exp-baseline}

\begin{figure*}[htbp]
\centering
     \begin{subfigure}[b]{0.24\textwidth}
         \centering
         \includegraphics[width=\textwidth]{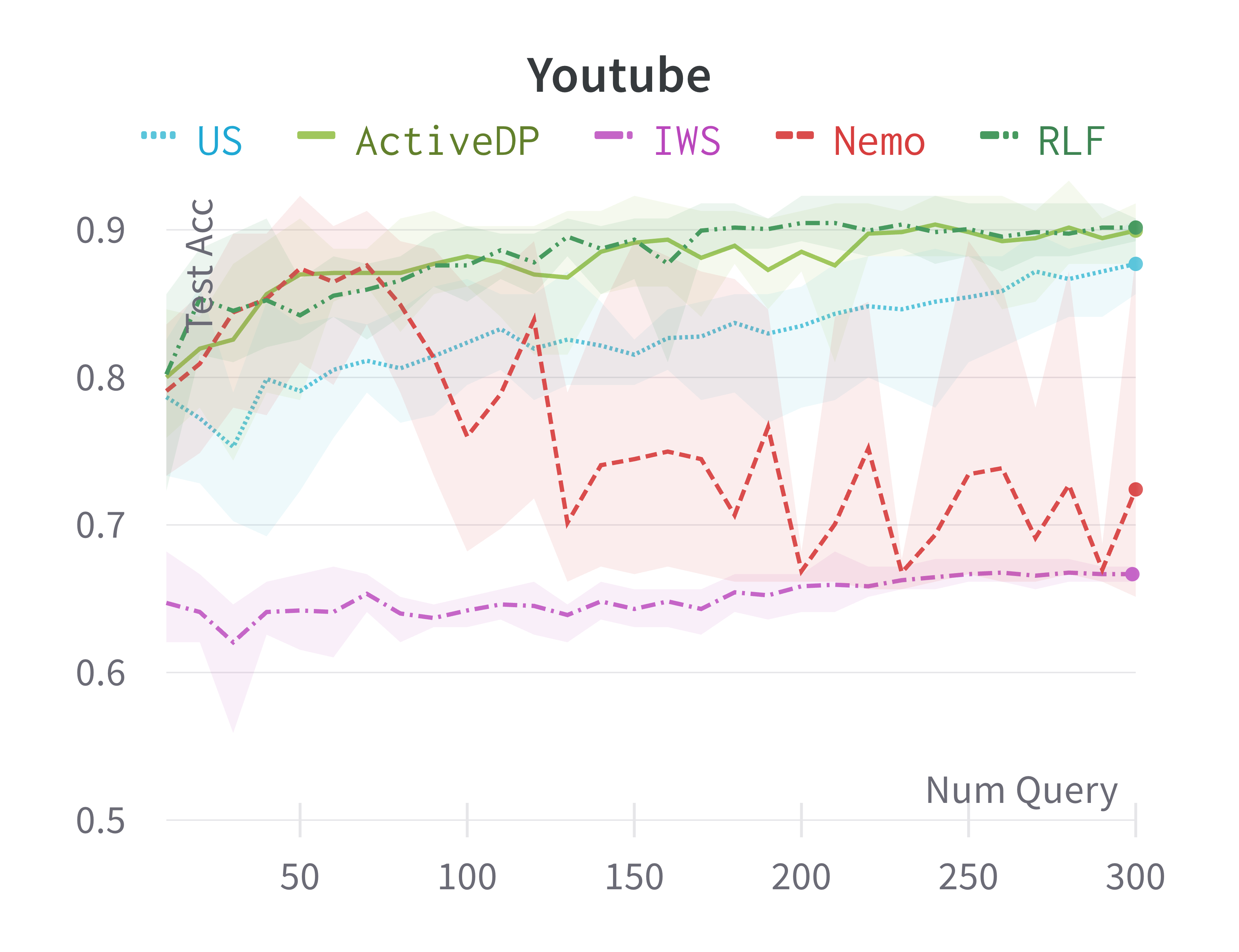}
     \end{subfigure}
     \hfill
     \begin{subfigure}[b]{0.24\textwidth}
         \centering
         \includegraphics[width=\textwidth]{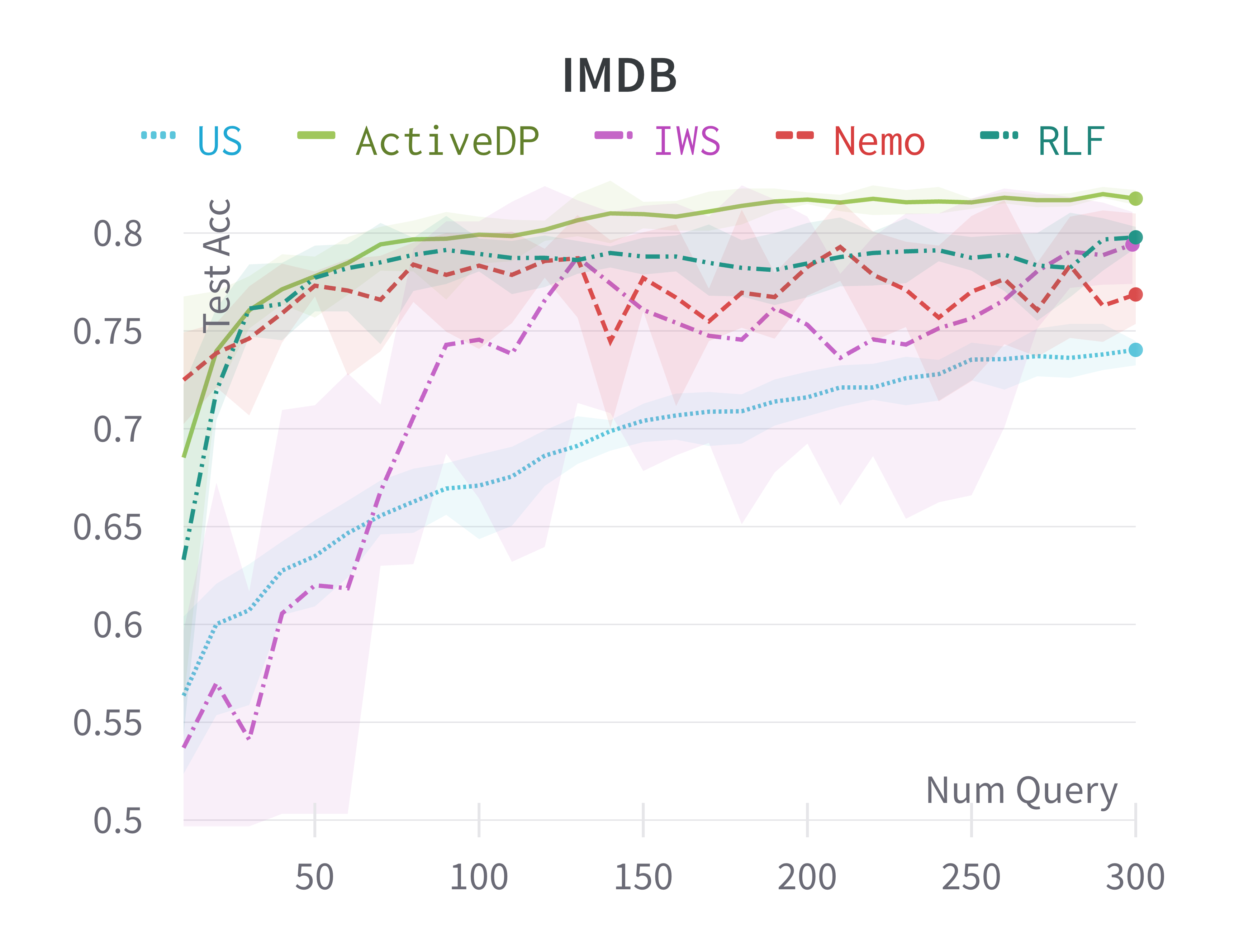}
     \end{subfigure}
     \hfill
     \begin{subfigure}[b]{0.24\textwidth}
         \centering
         \includegraphics[width=\textwidth]{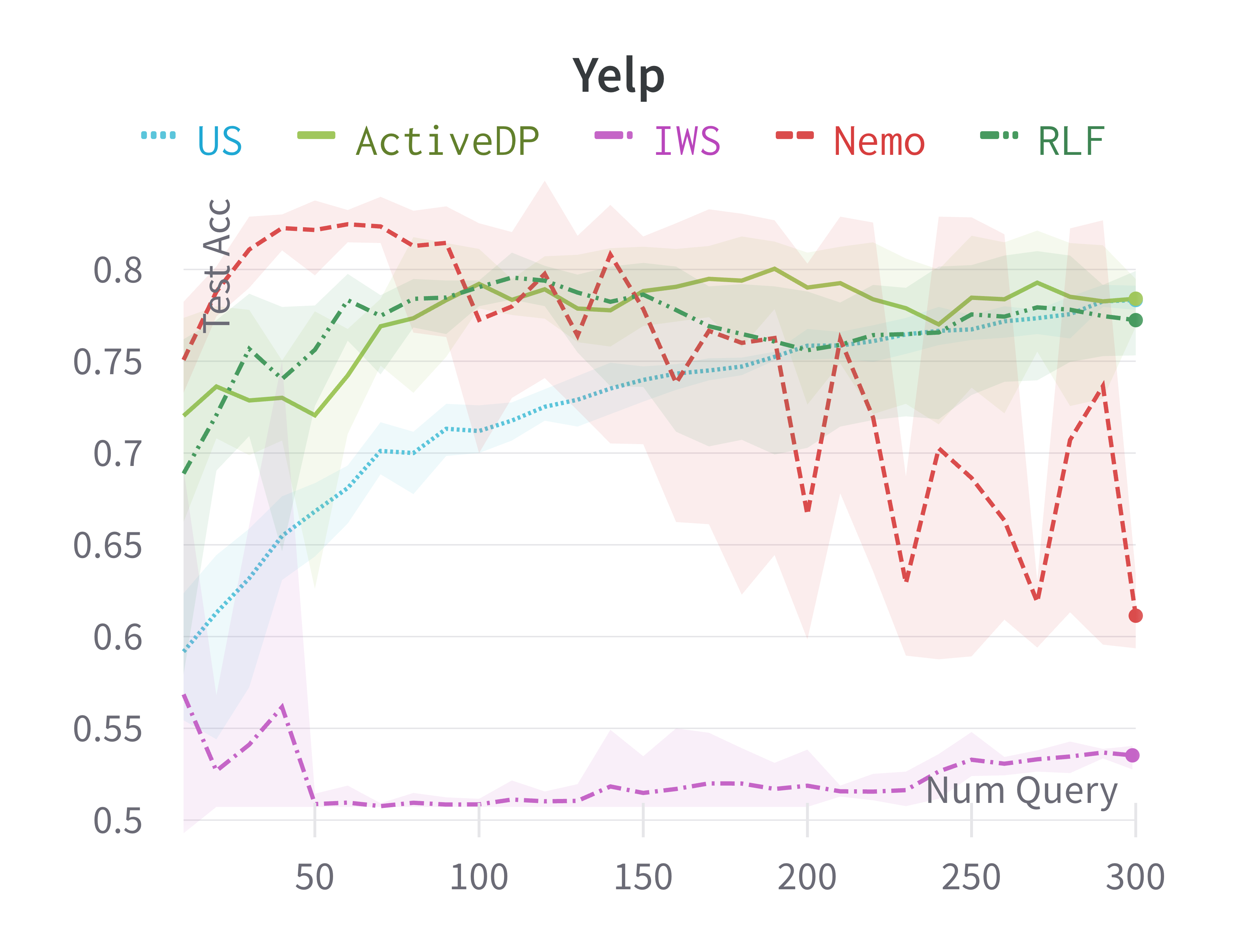}
     \end{subfigure}
     \hfill
     \begin{subfigure}[b]{0.24\textwidth}
         \centering
         \includegraphics[width=\textwidth]{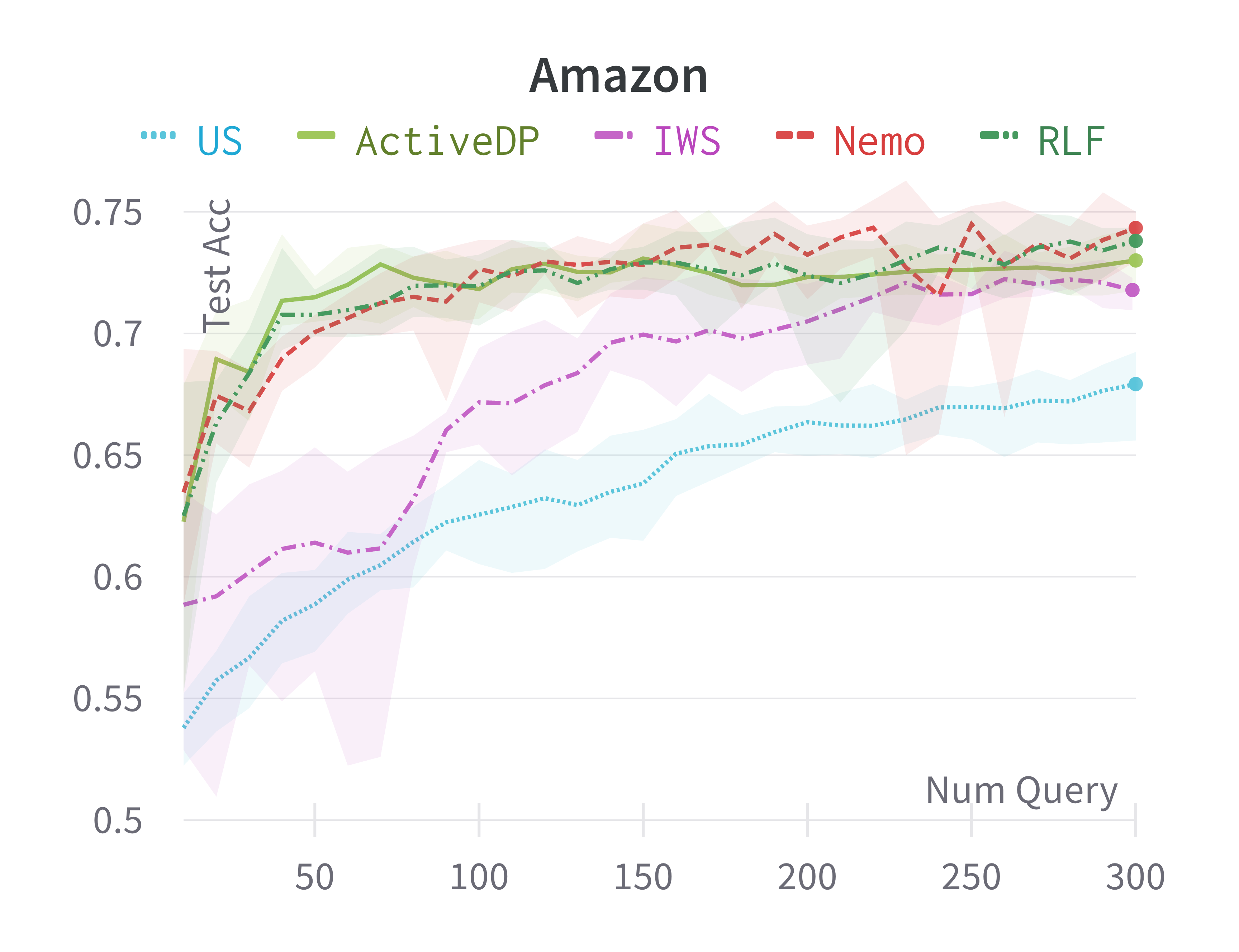}
     \end{subfigure}

     \begin{subfigure}[b]{0.24\textwidth}
         \centering
         \includegraphics[width=\textwidth]{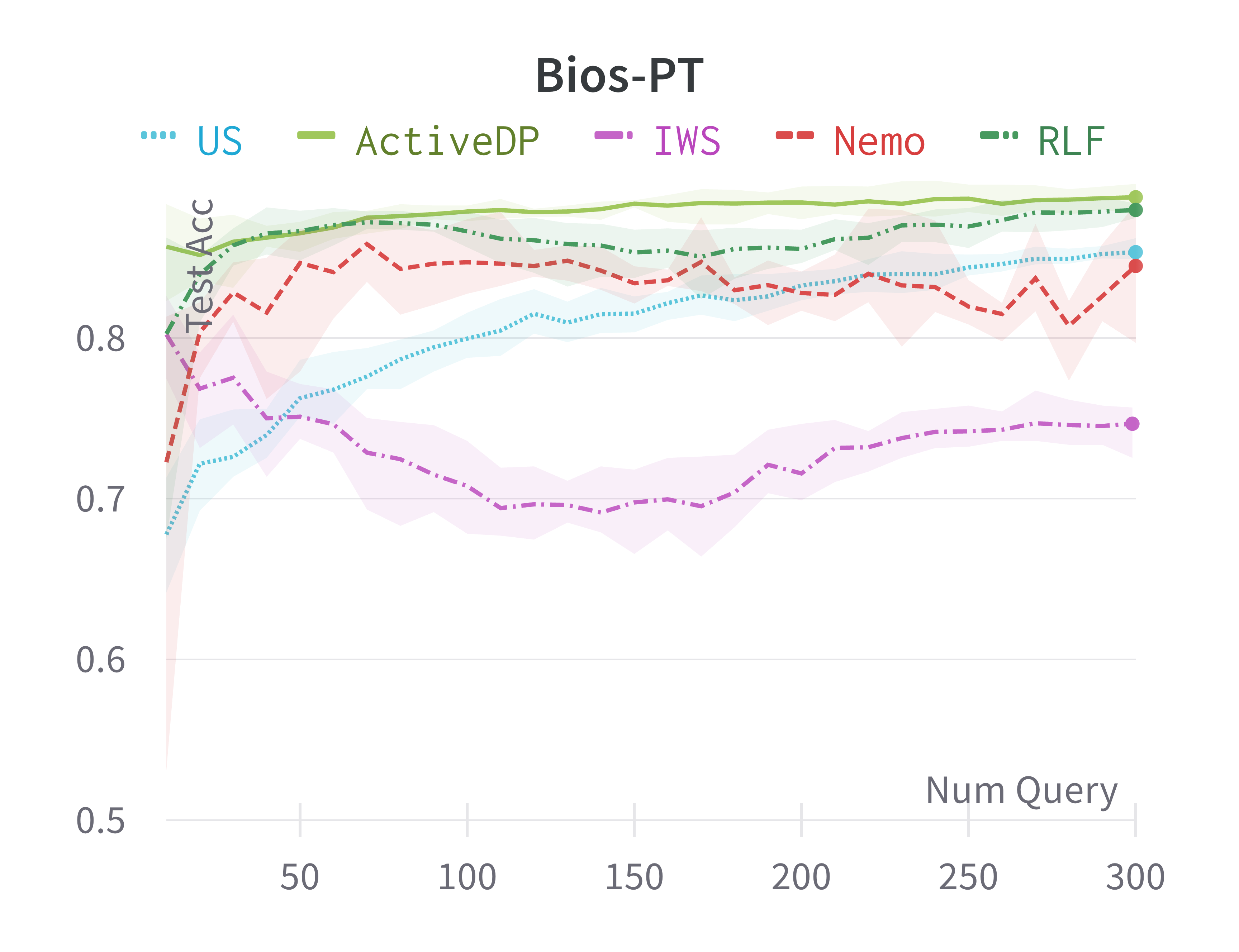}
     \end{subfigure}
     \hfill
     \begin{subfigure}[b]{0.24\textwidth}
         \centering
         \includegraphics[width=\textwidth]{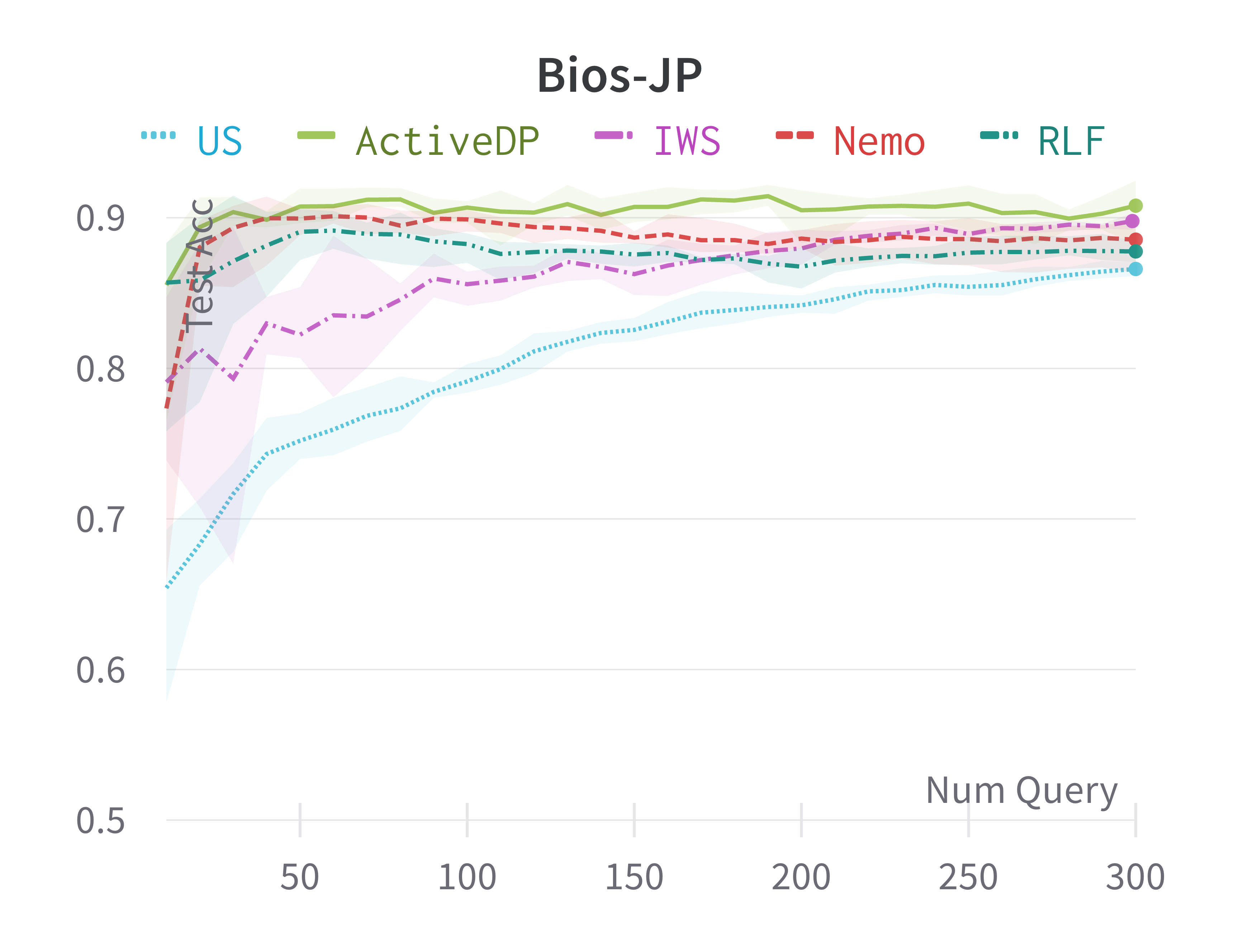}
     \end{subfigure}
     \hfill
     \begin{subfigure}[b]{0.24\textwidth}
         \centering
         \includegraphics[width=\textwidth]{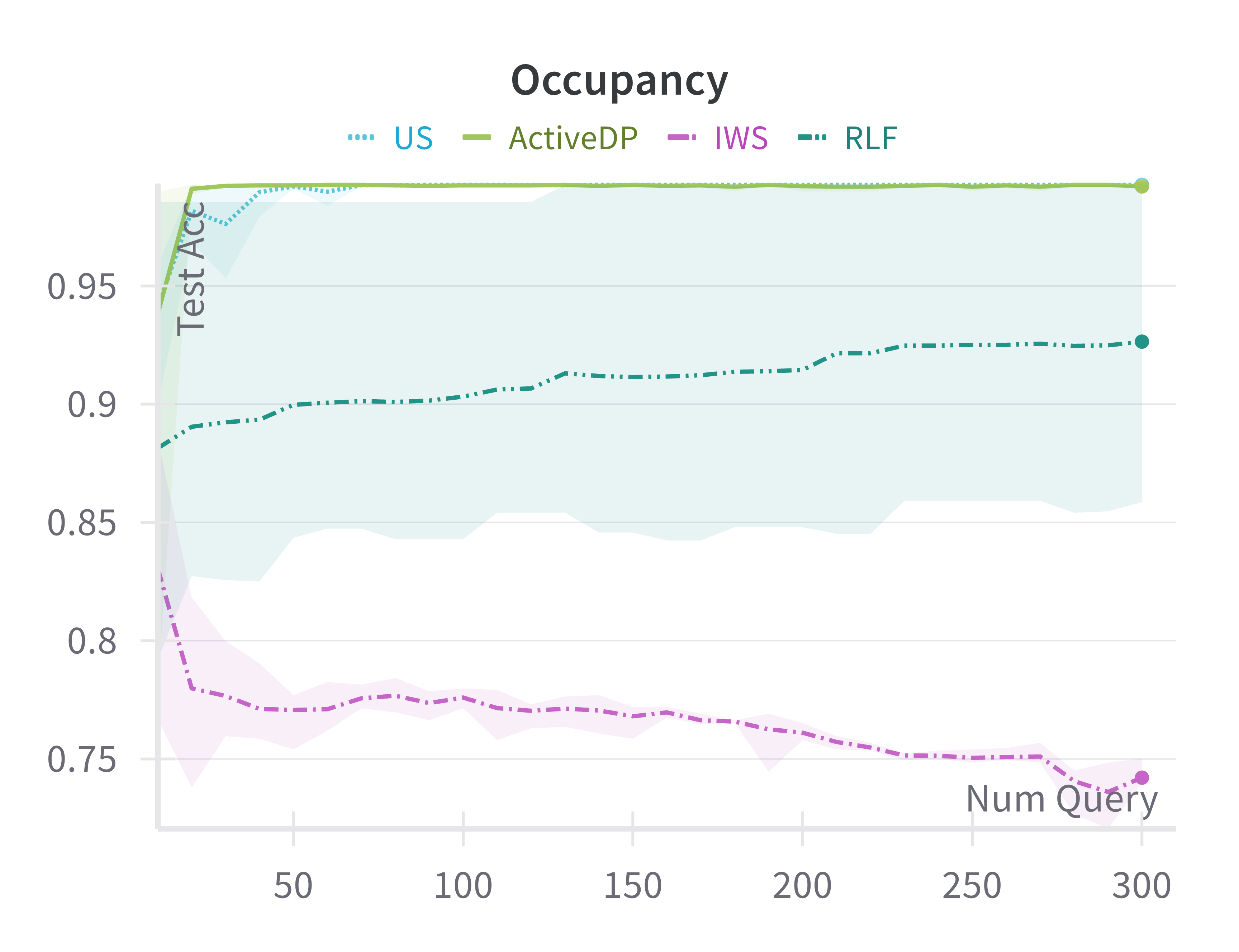}
     \end{subfigure}
     \hfill
     \begin{subfigure}[b]{0.24\textwidth}
         \centering
         \includegraphics[width=\textwidth]{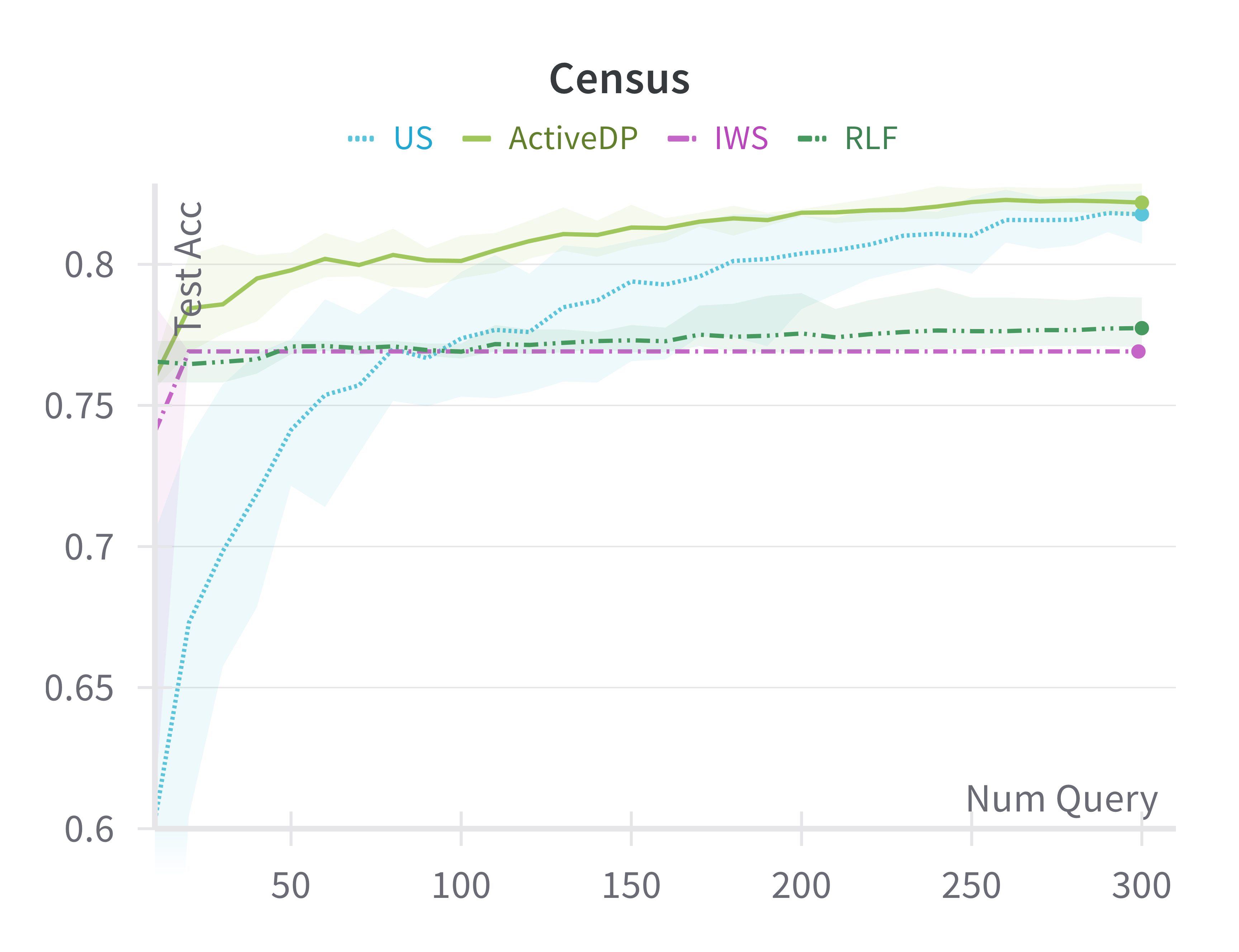}
     \end{subfigure}

    \caption{End-to-end Performance comparison between ActiveDP and Baseline Methods.}
    \label{fig:baseline}
\end{figure*}



Figure \ref{fig:baseline} shows the performance curve of ActiveDP and other baseline methods. The X-axis indicates the number of queries given to the simulated user, and the Y-axis shows the downstream model's test set classification accuracy. Across all datasets, ActiveDP improves the downstream model's test set accuracy by an average of 4.4\% compared to Nemo, 13.5\% compared to IWS, 2.6\% compared to Revising LF, and 6.5\% compared to uncertainty sampling. 

Looking into the labelling process in detail, when the label budget is small, ActiveDP, Nemo and Revising LF outperform uncertainty sampling and IWS in general. These three methods leverage data programming to efficiently label large datasets, giving them advantages at the early stage of the labelling process. On the other hand, uncertain sampling can only use a small labelled subset of data to train the downstream model, making it perform worse than data programming methods. IWS, although following the data programming paradigm, does not perform well in the early steps. This is likely because the system fails to provide good candidate LFs for human experts to verify when the labelled data is scarce. It underperforms Nemo and ActiveDP, which ask human experts to provide LFs instead. 


When the label budget increases, the performance of uncertain sampling and ActiveDP improves steadily, as they both train an active learning model which benefits from more labelled data. However, for Nemo and IWS, we observe a performance drop in some datasets. These methods only use label functions for prediction. As they omit fine-grained instance labels for supervision, their performance will be surpassed by active learning paradigms when the label budget is sufficient. Further, as these methods do not apply label function selection or only use a simple accuracy-based filter, their performance will be impeded when some harmful LFs are used to train the label model. 


For tabular datasets, the performance of uncertain sampling and ActiveDP improves with more labelling budget, the performance of Revising LF grows slowly, and the performance of IWS does not improve. This is because the LFs for tabular datasets are based on decision stumps, which can only provide coarse-grained supervision to the data. Thus developing more LFs does not significantly improve the data quality. On the other hand, these tabular datasets are relatively easy to classify using a small labelled dataset, thus the performance of uncertain sampling exceeds IWS and Revising LF after around 100 queries. By combining active learning and data programming, ActiveDP maintains good performance with only a few queries, and similar to uncertain sampling, it has steady performance improvement with increased labelling budgets.

Note that while the performance of Revising LF is close to ActiveDP in some textual datasets, the performance gaps in tabular datasets is significant. This is because the performance of Revising LF benefits from both aggregating more LFs and revising the LFs with user feedback. In textual datasets, both factors contribute to its performance improvement. However, in tabular datasets, the benefit of aggregating more LFs diminishes due to the reason explained above, leaving the effect of revising LFs to be dominant. As the LFs are only revised on queried instances, the performance improvement for Revising LF is much slower than ActiveDP or uncertain sampling, which trains an active learning model to predict a larger fraction of data. 

In summary, ActiveDP has good performance under both small and large labelling budget scenarios, and its performance improves steadily with more labels, making it a more general and effective solution in combining data programming and active learning compared to previous works.

\subsection{Comparative Studies} \label{sec:exp-comp}

\subsubsection{Ablation study.}\label{sec:exp-ablation}
\begin{table}[htbp]
\footnotesize
\centering
\caption{Performance of Ablated Versions of ActiveDP. }\label{tab:ablation}
\vspace{-0.1cm}
\resizebox{\columnwidth}{!}{
\begin{tabular}{c c c c c c c c c}
\toprule[1.5pt]
\multirow{2}{*}{\textbf{Method}} & \multicolumn{6}{c}{\textbf{Dataset}} \\
\cline{2-9}
& \textbf{Youtube} & \textbf{IMDB} & \textbf{Yelp} & \textbf{Amazon} & \textbf{Bios-PT} & \textbf{Bios-JP} & \textbf{Occupancy} & \textbf{Census} \\
\hline
Baseline & 0.8130 & 0.7826 & 0.5776 & 0.7096 & 0.8527 & 0.8892 & 0.8881 & 0.7651\\
LabelPick & 0.8598 & 0.7850 & 0.6564 & 0.7176 & 0.8572 & 0.9027 & 0.8881 & 0.7652 \\
ConFusion & 0.8708 & 0.7843 & 0.7538 & 0.7182 & 0.8671 & 0.8902 & 0.9906 & 0.8060\\
ActiveDP & 0.8894 & 0.8016 & 0.7834 & 0.7198 & 0.8785 & 0.9078 & 0.9905 & 0.8096\\
\bottomrule[1.5pt]
\end{tabular}}
\end{table}
We conduct ablation studies to investigate the benefit of the ConFusion and LabelPick methods. 
Table \ref{tab:ablation} evaluates the performance of the downstream model using different ablated versions of ActiveDP. The baseline method uses all user-returned LFs to train the label model; the LabelPick and ConFusion approach only apply the LF selection and confidence-based label aggregation techniques, respectively, and the ActiveDP method combines the two techniques. Compared to the baseline method, the LabelPick strategy improves the test set accuracy on average by 1.9\%, and the ConFusion strategy improves the test accuracy on average by 5.0\%. ActiveDP, combining both strategies, improves the test accuracy on average by 6.3\%, showing that both strategies benefit the performance of ActiveDP.

\subsubsection{Sampler Performance.}\label{sec:exp-sampler}
\begin{table}[htbp]
\footnotesize
\centering
\caption{Performance of ActiveDP with different sample selectors.}\label{tab:sampler}
\vspace{-0.1cm}
\resizebox{\columnwidth}{!}{
\begin{tabular}{c c c c c c c c c}
\toprule[1.5pt]
\multirow{2}{*}{\textbf{Sampler}} & \multicolumn{6}{c}{\textbf{Dataset}} \\
\cline{2-9}
& \textbf{Youtube} & \textbf{IMDB} & \textbf{Yelp} & \textbf{Amazon} & \textbf{Bios-PT} & \textbf{Bios-JP} & \textbf{Occupancy} & \textbf{Census} \\
\hline
Passive & 0.8699 & 0.7992 & 0.7607 & 0.7263 & 0.8692 & 0.9030 & 0.9823 & 0.7919\\

US & 0.8746 & 0.7997 & 0.7660 & 0.7212 & 0.8703 & 0.9058 & 0.9889 & 0.7959 \\

LAL & 0.8858 & 0.7889 & 0.7596 & 0.7242 & 0.8657 & 0.9002 & 0.9852 & 0.8033 \\

SEU & 0.8701 & 0.7566 & 0.7632 & 0.7259 & 0.8682 & 0.9023 & 0.9837 & 0.7927 \\

ADP & 0.8894 & 0.8016 & 0.7834 & 0.7198 & 0.8785 & 0.9078 & 0.9905 & 0.8096 \\
\bottomrule[1.5pt]
\end{tabular}}
\end{table}

To test the sensitivity of ActiveDP to the sample selector choice, we evaluated ActiveDP using different sample selection strategies: passive sampling, uncertainty sampling (US) \cite{lewis1995sequential}, learning active learning (LAL) \cite{konyushkova2017learning}, select by expected utility (SEU) \cite{hsieh2022nemo}, and the ADP sampler proposed by us. For the samplers designed for active learning (US, LAL), we use the implementation in the AliPy \cite{TLHalipy} package with default parameters.

Table \ref{tab:sampler} presents the downstream model performance using different samplers. Our ADP sampler outperforms other evaluated samplers on 7 out of 8 datasets. The ADP sampler is specifically designed to consider both active learning and data programming, making it a good fit for our ActiveDP framework.



\begin{table}[htbp]
\footnotesize
\centering
\caption{Performance of ActiveDP with different simulated Label Noise Rates.}\label{tab:label-noise}
\vspace{-0.1cm}
\resizebox{\columnwidth}{!}{
\begin{tabular}{c c c c c c c c c}
\toprule[1.5pt]
\multirow{2}{*}{\textbf{Label Noise}} & \multicolumn{6}{c}{\textbf{Dataset}} \\
\cline{2-9}
& \textbf{Youtube} & \textbf{IMDB} & \textbf{Yelp} & \textbf{Amazon} & \textbf{Bios-PT} & \textbf{Bios-JP} & \textbf{Occupancy} & \textbf{Census} \\
\hline
0\% & 0.8894 & 0.8016 & 0.7834 & 0.7198 & 0.8785 & 0.9078 & 0.9905 & 0.7948\\

5\% & 0.8671 & 0.7902 & 0.7449 & 0.7172 & 0.8692 & 0.9049 & 0.9905 & 0.7959\\

10\% & 0.8628 & 0.7846 & 0.7303 & 0.7091 & 0.8692 & 0.9036 & 0.9888 & 0.7922\\

15\% & 0.8474 & 0.7793 & 0.6999 & 0.7040 & 0.8583 & 0.9002 & 0.9884 & 0.7756\\
\bottomrule[1.5pt]
\end{tabular}}
\end{table}
\subsubsection{Label Noise}
 Label noise occurs when the LF has accuracy above the threshold but misfires on the corresponding query instances. We simulate label noise by randomly selecting a certain fraction of query instances and letting the simulated user generate LFs for the flipped label. The LF generation process is identical to the process described in Section \ref{sec:simulated-user}, except that as the label is reversed, the generated LFs will not be accurate on the query instance (their accuracy on the training set is still above the threshold of 0.6). As ActiveDP applies the LFs on related query instances to generate labelled subsets, such label noise will deteriorate the performance of the active learning model in ActiveDP. 

Table \ref{tab:label-noise} illustrates the performance under various label noise levels. While injecting label noise deteriorates the performance of ActiveDP, the performance degradation is not significant with a low label noise level. The average test set accuracy degradation is 1.1\% with 5\% label noise, 1.6\% with 10\% label noise, and 2.7\% with 15\% label noise.

\section{Conclusions} \label{sec:conclusion}
In this paper, we propose ActiveDP, a novel interactive framework that combines active learning and data programming to automatically generate labels for training ML models. ActiveDP selects a subset of helpful LFs for training the label model and leverages active learning to improve data quality. Extensive experiments show that ActiveDP outperforms previous weak supervision approaches and performs well under different labelling budgets.

\clearpage

\bibliographystyle{ACM-Reference-Format}
\bibliography{reference}

\end{document}